\documentclass[journal]{IEEEtran}

\title{Skin3D: Detection and Longitudinal Tracking \\ of Pigmented Skin Lesions in 3D Total-Body Textured Meshes}
\author{Mengliu Zhao$^*$\thanks{$^*$Authors contributed equally (joint first authors)}, Jeremy Kawahara$^*$, Kumar Abhishek, Sajjad Shamanian, and Ghassan Hamarneh\thanks{Corresponding author: hamarneh@sfu.ca (Ghassan Hamarneh)}
\thanks{Published in \emph{Medical Image Analysis} (2021)}
\thanks{Digital Object Identifier: \url{https://doi.org/10.1016/j.media.2021.102329}
}
\\
Medical Image Analysis Lab, School of Computing Science\\
Simon Fraser University, Canada}

\usepackage{multirow}
\usepackage[numbers]{natbib}

\usepackage{amssymb}
\usepackage{latexsym}

\usepackage[outdir=./]{epstopdf}

\usepackage[hyphens,spaces,obeyspaces]{url}
\usepackage{caption}
\usepackage{subcaption}
\usepackage{tabularx}
\usepackage{booktabs} 
\usepackage[bookmarks=false]{hyperref}

\hypersetup{pdfauthor=author}
\usepackage{mathtools}

\usepackage[automake,stylemods=longbooktabs,acronym,nonumberlist,sort=def]{glossaries-extra}
\setglossarystyle{superheaderborder}

\newglossary[nbg]{notation}{nbr}{nbn}{}
\glsdisablehyper

\makeglossaries
\newglossaryentry{mesh}{
    type=notation,
    name=\ensuremath{\mathcal{M}},
    description={A 3D mesh}
}

\newglossaryentry{vertices}{
    type=notation,
    name=\ensuremath{V},
    description={A set of 3D vertices}
}

\newglossaryentry{vertex}{
    type=notation,
    name=\ensuremath{v},
    description={A single 3D vertex}
}

\newglossaryentry{faces}{
    type=notation,
    name=\ensuremath{F},
    description={A set of 3D faces}
}

\newglossaryentry{face}{
    type=notation,
    name=\ensuremath{\xi},
    description={A single 3D face}
}

\newglossaryentry{uvcoords}{
    type=notation,
    name=\ensuremath{U},
    description={A set of UV texture coordinates}
}

\newglossaryentry{uvcoord}{
    type=notation,
    name=\ensuremath{u},
    description={A single UV texture coordinates}
}

\newglossaryentry{textureImage}{
    type=notation,
    name=\ensuremath{I},
    description={A 2D texture image}
}

\newglossaryentry{rcnn}{
    type=notation,
    name=\ensuremath{\phi},
    description={Faster R-CNN}
}

\newglossaryentry{parameters}{
    type=notation,
    name=\ensuremath{\theta},
    description={The learned parameters of a Faster R-CNN}
}

\newglossaryentry{lesions2d}{
    type=notation,
    name=\ensuremath{Y},
    description={Set of 2D lesion bounding boxes}
}

\newglossaryentry{lesion2d}{
    type=notation,
    name=\ensuremath{y},
    description={2D lesion bounding box}
}

\newglossaryentry{predLesions2d}{
    type=notation,
    name=\ensuremath{\hat{Y}},
    description={Set of predicted 2D bounding boxes}
}

\newglossaryentry{predLesion2d}{
    type=notation,
    name=\ensuremath{\hat{y}},
    description={Predicted 2D lesion bounding box}
}

\newglossaryentry{predLesions3d}{
    type=notation,
    name=\ensuremath{\hat{Z}},
    description={Set of predicted 3D lesion center coordinates}
}

\newglossaryentry{predLesion3d}{
    type=notation,
    name=\ensuremath{\hat{z}},
    description={Predicted 3D lesion center coordinates}
}

\newglossaryentry{templateVertices}{
    type=notation,
    name=\ensuremath{A},
    description={The vertices of the template mesh}
}

\newglossaryentry{reconVertices}{
    type=notation,
    name=\ensuremath{R},
    description={The set of 3D reconstructed vertices}
}

\newglossaryentry{reconVertex}{
    type=notation,
    name=\ensuremath{r},
    description={A single 3D reconstructed vertex}
}

\newglossaryentry{3dmatching}{
type=notation,
name=\ensuremath{\pi},
description={Matching matrix of 3D vertices}
}

\newglossaryentry{nodedistance}{
type=notation,
name=\ensuremath{g},
description={Distance measure between a pair of vertices}
}

\newglossaryentry{node_geodesic}{
type=notation,
name=\ensuremath{d_g},
description={Geodesic distance calculated on a 3D mesh}
}

\newglossaryentry{3d_match_loss}{
type=notation,
name=\ensuremath{\mathcal{L}},
description={3D matching loss}
}

\newglossaryentry{unary}{
type=notation,
name=\ensuremath{\mathcal{U}},
description={Unary term of 3D matching loss}
}

\newglossaryentry{binary}{
type=notation,
name=\ensuremath{\mathcal{B}},
description={Binary term of 3D matching loss}
}

\newglossaryentry{3dregistration}{
type=notation,
name=\ensuremath{\mathcal{V}},
description={Function to determine corresponding vertices}
}

\newglossaryentry{dummynode}{
    type=notation,
    name=\ensuremath{z_\Phi},
    description={A dummy node in graph matching}
}

\newglossaryentry{dummy_u_cost}{
    type=notation,
    name=\ensuremath{C_u^{\Phi}},
    description={Dummy node constant unary cost}
}

\newglossaryentry{dummy_b_cost}{
    type=notation,
    name=\ensuremath{C_b^{\Phi}},
    description={Dummy node constant binary cost}
}

\newglossaryentry{realNumbers}{
    type=notation,
    name=\ensuremath{\mathbb{R}},
    description={The set of real numbers}
}

\newcommand{\numberBodyTexLesionsAnnotated}{26,507}

\newcommand{\annotatedBodyTexAvgHeight}{22.95}
\newcommand{\annotatedBodyTexAvgWidth}{22.07}

\newcommand{\numberBodyTexScansAnnotated}{218}

\newcommand{\numberBodyTexTrainSubjects}{120}
\newcommand{\numberBodyTexTrainScans}{128}

\newcommand{\numberBodyTexValidSubjects}{40}

\newcommand{\numberBodyTexTestSubjects}{40}

\newcommand{\numberBodyTexLongScans}{10}

\usepackage{amsmath}

\DeclareMathOperator*{\argmin}{argmin}

\def\ie{i.e.,~}
\def\eg{e.g.,~}

\def\fasterrcnn{Faster R-CNN}

\begin{document}

\IEEEoverridecommandlockouts
\IEEEpubid{\begin{minipage}[t]{\textwidth}\ \\[10pt]
        \centering\footnotesize{\copyright 2021. This manuscript version is made available under the CC-BY-NC-ND 4.0 license \url{https://creativecommons.org/licenses/by-nc-nd/4.0/}}
\end{minipage}} 

\maketitle






\begin{abstract}
We present an automated approach to detect and longitudinally track skin lesions on 3D total-body skin surface scans. The acquired 3D mesh of the subject is unwrapped to a 2D texture image, where a trained objected detection model, \fasterrcnn, localizes the lesions within the 2D domain. These detected skin lesions are mapped back to the 3D surface of the subject and, for subjects imaged multiple times, we construct a graph-based matching procedure to longitudinally track lesions that considers the anatomical correspondences among pairs of meshes and the geodesic proximity of corresponding lesions and the inter-lesion geodesic distances.

We evaluated the proposed approach using 3DBodyTex, a publicly available dataset composed of 3D scans imaging the coloured skin (textured meshes) of 200 human subjects. We manually annotated locations that appeared to the human eye to contain a pigmented skin lesion as well as tracked a subset of lesions occurring on the same subject imaged in different poses. Our results, when compared to three human annotators, suggest that the trained \fasterrcnn~detects lesions at a similar performance level as the human annotators. Our lesion tracking algorithm achieves an average matching accuracy of 88\% on a set of detected corresponding pairs of prominent lesions of  subjects imaged in different poses, and an average longitudinal accuracy of 71\% when encompassing additional errors due to lesion detection. As there currently is no other large-scale publicly available dataset of 3D total-body skin lesions, we publicly release over 25,000 3DBodyTex manual annotations, which we hope will further research on total-body skin lesion analysis.

\end{abstract}

\section{Introduction}
Skin diseases are the most common reasons for patients to visit general practitioners in studied populations~\citep{schofield2011skin}. In the United States, melanoma-related cancer is estimated to increase by 100 thousand cases and cause more than 7000 deaths in 2021 ~\citep{cancerfact2021}, while the accessibility and wait time to see a dermatologist has become a concern~\citep{feng2018comparison, creadore2021insurance}. Leveraging the ability of artificial intelligence, especially in assisting the diagnosis of whole-body melanoma-related skin cancer, may improve the early diagnosis efficiency and improve patients' outcomes~\citep{shellenberger2016melanoma}. 

When diagnosing a lesion, considering a large region of the skin and changes to lesions over time may provide additional helpful context not available when only considering a localized lesion at a single point in time. For example, visualizing a wide region of the skin may allow for multiple nevi to be detected, which is an important risk factor for melanoma~\citep{Gandini2005}. When monitoring a lesion over time, a nevi that is changing is an additional melanoma risk factor~\citep{Abbasi2004}. As an estimated 30\% of melanomas evolve from existing nevi~\citep{Pampena2017}, imaging and tracking a lesion across time may lead to a better understanding of how a lesion may transform and allow for early prediction of melanoma~\citep{Sondermann2019}, reduce the number of excisions~\citep{tschandl2018sequential}, and improve the prognosis of certain melanoma risk groups~\citep{haenssle2010selection}. Further, clinical studies suggest a benefit to patient outcome through longitudinal total-body skin imaging. Monitoring high risk patients with follow-up visits using total-body photography has been found to be associated with a greater rate of detecting lower-risk melanomas and overall survival~\citep{Strunck2020}. Combining digital total-body photography and dermoscopy to monitor patients over time has been found to enable early detection of melanoma with a low excision rate~\citep{Salerni2012}.

While total-body skin imaging and longitudinal monitoring of skin lesions shows promise to improve melanoma detection, the amount of manual intervention required by human dermatologists to monitor and track changes of multiple lesions over the entire body is significant. With interest in collecting 3D reconstructions of the human body surface growing among dermatologists~\citep{Rayner2018,Primiero2019}, we expect computer-aided approaches capable of detecting and tracking lesions over the entire body to reduce manual efforts and, in addition, serve as a system to flag high risk lesions for review by human dermatologists.

\subsection{Tracking and Detecting Multiple Lesions}
The majority of computer-aided approaches to analyze pigmented skin lesions rely on static 2D colour images showing a single lesion, with limited works focusing on total-body photography or tracking lesions across time~\citep{Celebi2019,Pathan2018}. 
Rather than focusing on a single lesion, a \emph{wide-area} (\eg torso) region of the body can be imaged, where multiple lesions are visible. Using wide-area images, \citet{Lee2005} proposed an unsupervised approach to segment and count moles from 2D colour images of the back torso, but did not explore tracking lesions across time or incorporate total-body images.

Other computer-aided approaches explore changes to a \emph{single lesion} across time. For example, to measure changes within a lesion across time, \citet{Navarro2019} proposed to register two dermoscopy images of the same lesion taken at different times and combined an automatic lesion segmentation approach to evaluate how a skin lesion evolves. Their approach assumes the corresponding lesions are known and the images are acquired using dermoscopy, which provide a more controlled field-of-view and scale than non-dermoscopy images.

Combining both wide-area images and lesions imaged over time, \citet{McGregor1998} registered (non-dermoscopy) photographs of human subjects by automatically detecting and selecting a matching subset of the lesions used to transform the images. Their study restricted the imaging to the torso regions and did not explore total-body imaging. \citet{Mirzaalian2016} proposed an approach to track skin lesions by detecting lesions on 2D wide-area skin images of the back of a body and matching lesions across images of the same subject using detected anatomical landmarks. \citet{li2016skin} proposed a data augmentation strategy to synthesize large body images by blending skin lesion images onto images of back, legs, and chest using Poisson image editing, and then trained a fully convolutional neural network to perform lesion detection and a CNN to track lesions over time by estimating pixel-wise correspondence between a pair of images of the same body part. More recently, \citet{Soenksen2021} proposed a multi-stage workflow for detecting pigmented skin lesions from 2D wide-field images by first using a blob detection algorithm to extract all regions of interest in an image followed by CNN-based classification of these regions as suspicious or non-suspicious skin lesions or other classes (skin only, skin edge, or other background material).

Expanding the imaging field-of-view, \emph{total-body} photography allows for the entire skin (or nearly the entire skin) to be imaged. \citet{Korotkov2015} developed an image acquisition system for photogrammetry-based total-body scanning of the skin to automatically acquire overlapping images of the patient's skin. As the scanning system is in a carefully controlled environment, lesions can be automatically tracked across scans. \citet{Korotkov2019} further extended their system to improve mole matching among images, but did not explore 3D reconstruction of skin surfaces. 

The 2014 work by \citet{Bogo2014a} is similar to ours, where \citet{Bogo2014a} performed whole-body 3D imaging, registered body scans acquired at multiple times, detected candidate lesions using linear discriminant analysis, and tracked lesions based on the registered body locations. We differ in our approach in that we use recently developed deep-learning based approaches to both detect lesions and register meshes, where our lesion detection step does not require choosing a threshold to isolate the skin nor post-processing as done in~\citet{Bogo2014a}. For lesion tracking, \citet{Bogo2014a} used a common UV template to map anatomically corresponding locations to common texture elements within a texture image, while we perform our tracking on the 3D coordinates and consider geodesic distances between pairs of lesions. Finally, we note we use a larger dataset (200 vs. 22 unique subjects) with manually detected lesions (vs. artificially drawn lesions).

\subsection{Longitudinal Skin Datasets with Multiple Lesions}
When compared to the large body of literature using static, single lesion images to analyze skin lesions, there are relatively few skin works for tracking multiple lesions across time. This is likely due to the lack of large standardized annotated datasets of longitudinal images available for computer-aided research, prompting a call by clinicians to gather longitudinal skin lesion images to train both clinicians and computer algorithms~\citep{Sondermann2019}.

The publicly available FAUST dataset~\citep{Bogo2014} provides 300 3D body meshes, but lacks colour information about the surface of the body (\eg skin colour information). \citet{Saint2018} created the 3DBodyTex dataset, which consists of 400 high resolution, 3D texture scans of 100 male and 100 female subjects captured in two different poses. Some lesions are visible on the skin; however, there are no lesion annotations publicly available. To the best of our knowledge, there currently is no large scale longitudinal dataset imaging a wide region of the skin with multiple annotated lesions publicly available. This lack of annotated skin lesion data makes it challenging to train machine learning models that require large amounts of labeled data and to benchmark competing approaches. 

\subsection{Contributions}
Methodologically, we propose a novel approach to detect and track skin lesions across 3D total-body textured meshes, where the colours on the surface of the subject are unwrapped to yield a 2D texture image, a \fasterrcnn~model detects the lesions on the 2D texture images, and the detected lesions in 2D are mapped back to 3D coordinates. The detected lesions are tracked in 3D across pairs of meshes of the same subject using an optimization method that relies on deep learning-based anatomical correspondences and the geodesic distances between lesions. Our proposed approach has the potential for clinical utility in which a trained model detects the lesions on the 2D texture image, tracks the lesions across meshes in 3D, and displays the tracked lesions on the 3D mesh (or a 2D view of the 3D mesh) for the dermatologist to review, thereby reducing the initial manual effort required to densely annotate and track lesions over large skin regions. To the best of our knowledge, this is the largest evaluation of 3D total-body skin lesion detection and tracking on 3D textured meshes with skin colour information. Further, we make publicly available\protect\footnote{ \url{https://github.com/jeremykawahara/skin3d}} over 25,000 manual lesion annotations for 3DBodyTex.

\section{Methods}
\label{skin3d:sec:methods}
To detect lesions on the surface of 3D models and track lesions across time, we propose using 3D textured meshes, where the colours on the surface of the subject  are mapped to 2D texture images, lesions are detected on 2D texture images, and tracked in 3D for longitudinal monitoring. A summary of the mathematical notation used in this paper is provided in Table~\ref{tab:glossary}.

A scanned 3D object (\eg human subject) is represented as a 3D mesh $\gls{mesh}=(\gls{vertices}, \gls{faces}, \gls{uvcoords}, \gls{textureImage})$, with $N$ vertices $\gls{vertices} \in \gls{realNumbers}^{N \times 3}$ where a single vertex $\gls{vertex} \in \gls{vertices}$ describes a 3D spatial coordinate; $\gls{face} \in \gls{faces}$ is the 3D face set which encodes the edges among the vertices; and $\gls{uvcoords} \in \gls{realNumbers}^{N \times 2}$ are 2D UV texture coordinates that map vertices $\gls{vertices}$ to the texture image $\gls{textureImage} \in \gls{realNumbers}^{W \times H \times 3}$, where $W$ and $H$ indicate the width and height of the colour texture image. Given meshes of the same subject scanned at $T$ different points in time $\{\gls{mesh}_t\}_{t=1}^T$, our goal is to detect the skin lesions within each scan, and match the corresponding lesions across the scans.

\begin{center}
\captionsetup{type=table}
\captionof{table}{Summary of Notations}
\vspace{-2em}
\printglossaries
\label{tab:glossary}
\end{center}

\subsection{Detecting Lesions on 2D Texture Images}
\label{skin3d:sec:detectLesion}

To detect lesions, we rely on the 2D texture image \gls{textureImage}, and represent a single lesion with a bounding box $\gls{lesion2d} \in \gls{realNumbers}^4$ defined by 2D coordinates of two of its corners (2D coordinates of the top-left and bottom-right corners). We represent the bounding boxes of $M$ lesions within \gls{textureImage} as $\gls{lesions2d} \in \gls{realNumbers}^{M \times 4}$.

We represent a lesion with a 2D bounding box since our goal in this work is to detect (\ie localize and classify the existence of) the lesion from the background skin, which can be obtained via a bounding box, rather than acquiring a precise lesion segmentation. Precisely annotating the boundaries of the lesion is challenging due to the resolution of the texture images. Further, manually annotating 2D bounding boxes is far less time consuming for human annotators compared to 3D delineations.

Using the 2D texture image \gls{textureImage} and corresponding lesion 2D bounding box annotations \gls{lesions2d}, we trained a region convolutional neural network (\fasterrcnn) (details in Section~\ref{skin3d:sec:rcnn}) to predict 2D bounding box coordinates,
\begin{equation}
    \gls{predLesions2d} = \gls{rcnn} \left( \gls{textureImage}; \gls{parameters} \right)
    \label{skin3d:eqn:lesionDetector}
\end{equation}
where $\gls{rcnn}(\cdot)$ is the \fasterrcnn~trained to detect skin lesions with learned parameters \gls{parameters}; and \gls{predLesions2d} are the predicted 2D lesion bounding box coordinates within \gls{textureImage}. We use these 2D predicted bounding box lesion coordinates to measure the lesion detection performance between human and machine annotations (see results in Section~\ref{skin3d:sec:3dbodytex-static}). However, for visualization purposes, and for tracking lesions across time, we require the 2D lesion locations to be mapped back to 3D coordinates.

\begin{figure}[ht]
\centering
\includegraphics[width=\linewidth]{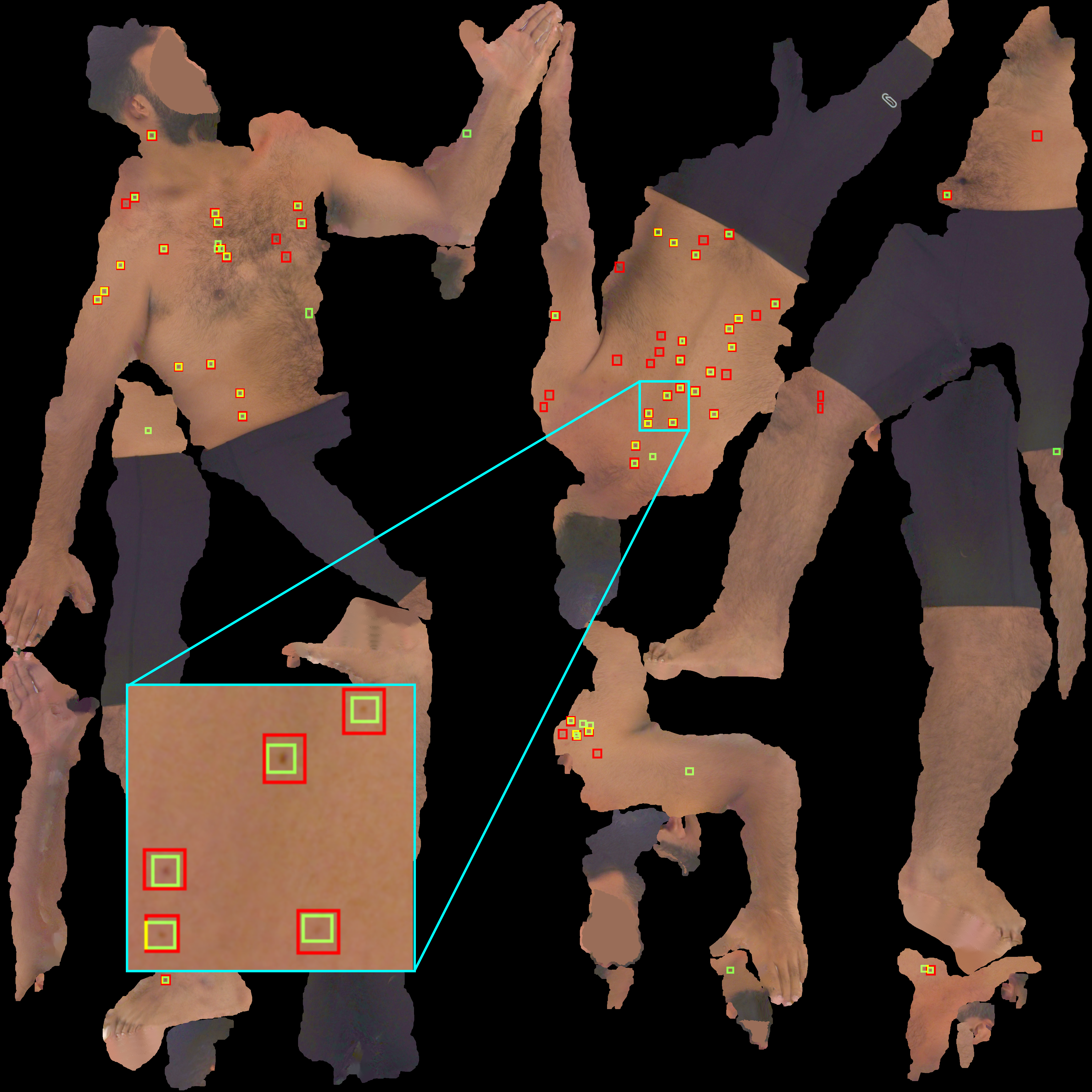}
\caption{A texture image with lesions detected manually (green box) and automatically (red box). The yellow borders indicate overlapping manual and automatic boxes. The blue box shows a closeup region of the texture image.}
\label{skin3d:fig:bodytex-detect2d}
\end{figure}

\subsection{Mapping the Detected 2D Lesions onto a 3D Mesh}
\begin{figure}[htb]
\centering
\begin{subfigure}[b]{0.50\linewidth}
\includegraphics[width=\linewidth]{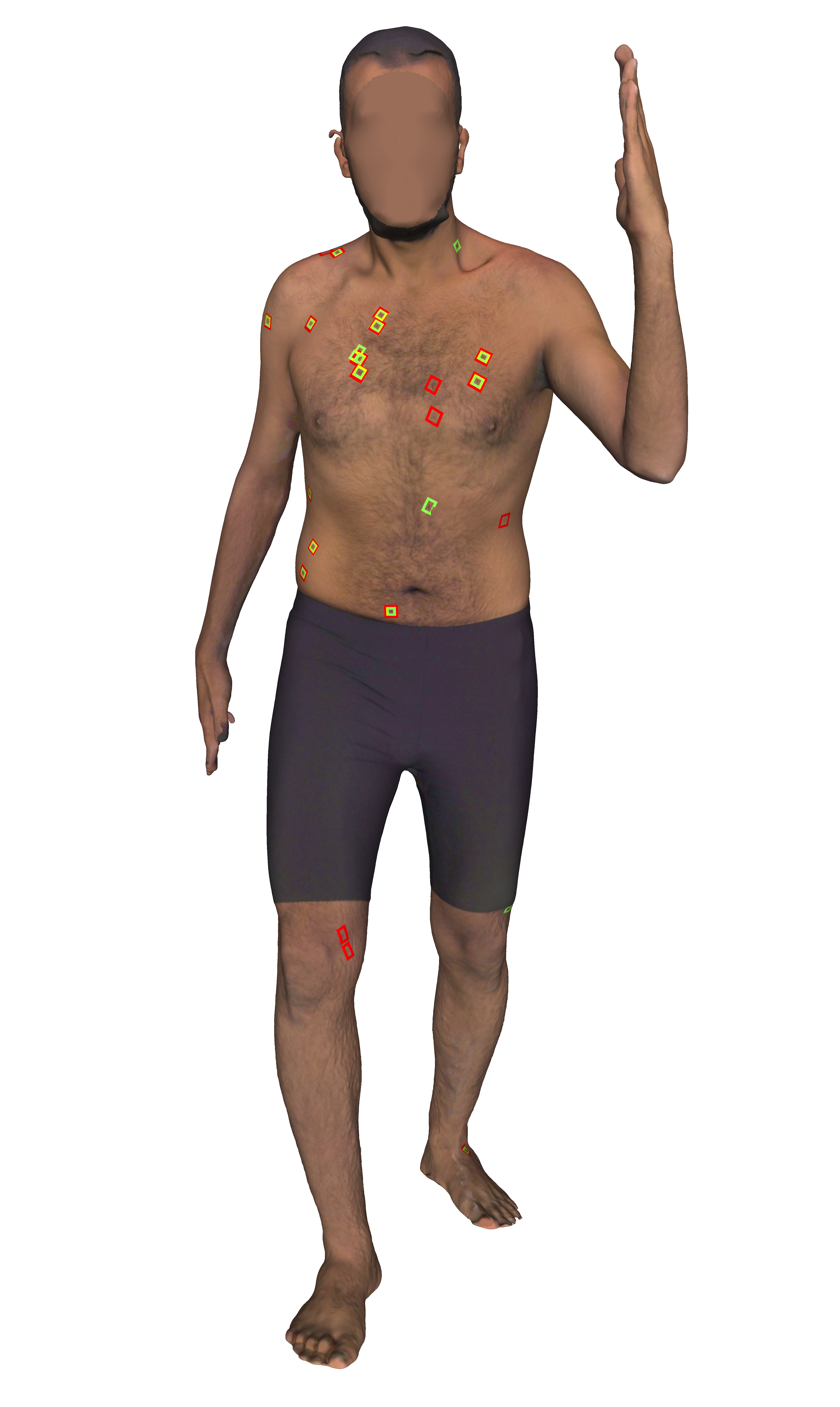}
\end{subfigure}
\begin{subfigure}[b]{0.47\linewidth}
\includegraphics[width=\linewidth]{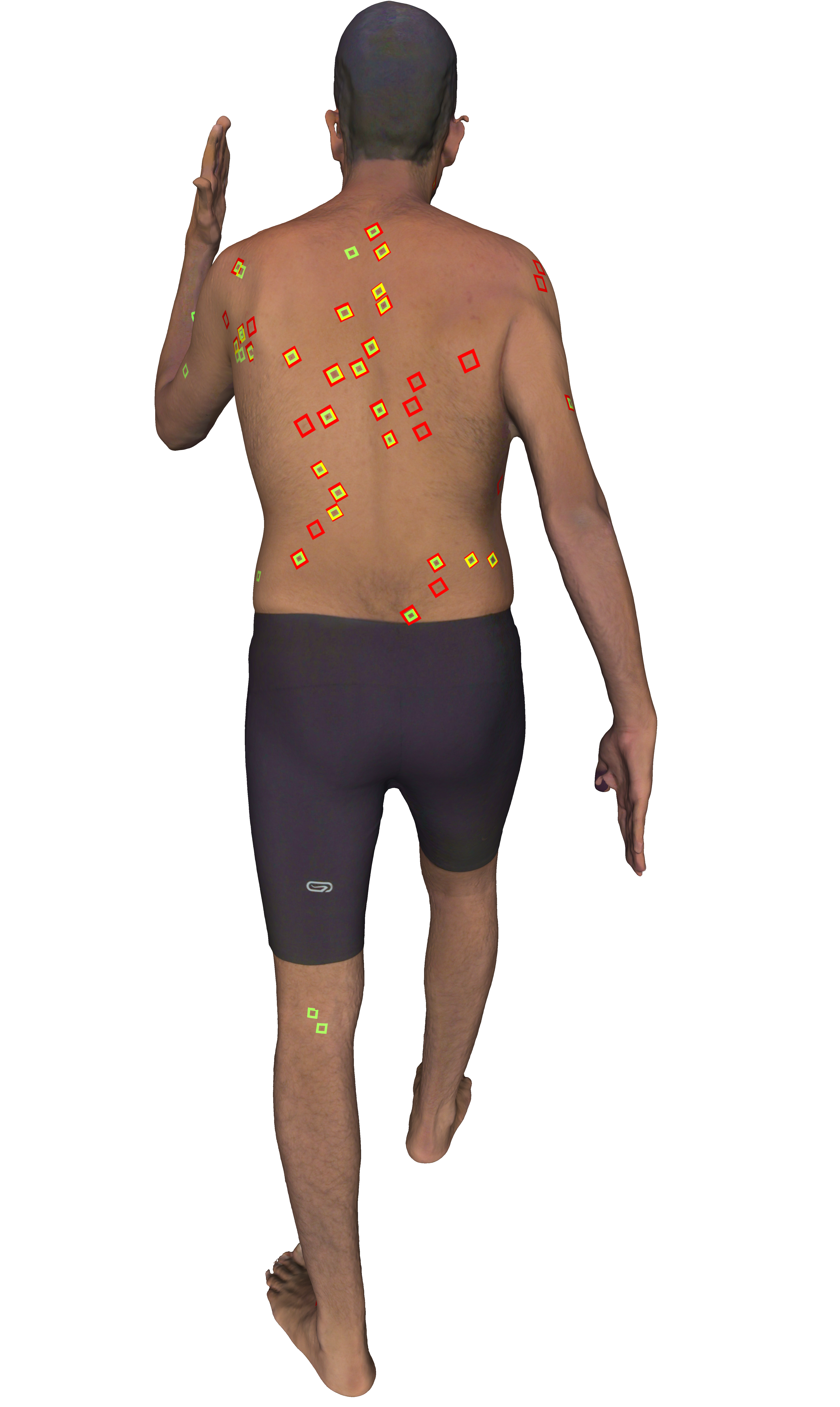}
\end{subfigure}
\caption{The front and back of a subject with manual and automatically annotated lesions after mapping the texture image in Fig.~\ref{skin3d:fig:bodytex-detect2d} onto a 3D body mesh.}
\label{skin3d:fig:bodytex-detect3d}
\end{figure}

To visualize the 2D bounding boxes on the 3D mesh, we modify the texture elements based on the coordinates of the lesion bounding boxes \gls{predLesions2d} in order to embed the bounding boxes within the original texture image \gls{textureImage}. The resulting lesion embedded texture image \gls{textureImage}$_{\gls{predLesions2d}}$ (Fig.~\ref{skin3d:fig:bodytex-detect2d}) replaces the original texture image \gls{textureImage} to show (Fig.~\ref{skin3d:fig:bodytex-detect3d}) the embedded lesion annotations and texture information on the 3D mesh $\gls{mesh}_{\gls{predLesions2d}} = (\gls{vertices}, F, \gls{uvcoords}, \gls{textureImage}_{\gls{predLesions2d}})$.

While bounding boxes are well suited for visualizing the localized lesions, we perform 3D analysis using the 3D positions of the lesions in order to better determine lesion correspondence across meshes, where the geodesic distances between lesions is computed on the 3D shape of the human body (discussed in Section~\ref{skin3d:sec:3dcoded}). We represent a lesion with the 3D vertex closest to the UV coordinates of the lesion's 2D center point. Specifically, given the $i$-th lesion \gls{predLesion2d}$^{(i)}$ detected within \gls{textureImage}, we find the index of the UV coordinates closest to the center point of the 2D bounding box,
\begin{equation}
    j^* = \argmin_{j \in \{1, \dots, N \}} L_1(\gls{uvcoord}^{(j)}, f(\gls{predLesion2d}^{(i)}))
    \label{skin3d:eqn:closestUV}
\end{equation}
where $f(\cdot)$ computes the 2D center of the lesion bounding box and converts these image coordinates to the UV domain; $\gls{uvcoord}^{(j)} \in U$ returns the $j$-th UV coordinates; and $L_1(\cdot)$ computes the $L_1$ distance between the mesh's UV coordinates and the predicted lesion's UV coordinates. As the indexes of the UV coordinates \gls{uvcoords} correspond to the indexes of the vertices \gls{vertices} (\ie the UV coordinates that correspond to \gls{vertex}$^{(j)}$ are $\gls{uvcoord}^{(j)}$), we obtain the mesh's 3D vertex that corresponds to the 2D lesion coordinates by indexing the $j^*$-th vertex, \gls{vertex}$^{(j^*)}$. Thus we map the $i$-th 2D lesion to the $j^{*}$-th 3D vertex within the mesh,
\begin{equation}
    \gls{predLesion2d}^{(i)} \longrightarrow \gls{vertex}^{(j^*)}
    \label{skin3d:eqn:mapped3d}
\end{equation}
where Eq.~\ref{skin3d:eqn:closestUV} is used to determine this mapping. We note that in Eq.~\ref{skin3d:eqn:closestUV} and Eq.~\ref{skin3d:eqn:mapped3d} we approximate the location of the lesion with the nearest mesh vertex. We apply this approximation to reduce the complexity of computing the geodesic distances across meshes (discussed in section~\ref{skin3d:sec:tracking}). 

\subsection{Anatomically Corresponding Vertices Across Scans}
\label{skin3d:sec:3dcoded}
\begin{figure}[ht]
\centering
\begin{subfigure}[b]{0.24\linewidth}
\includegraphics[width=\linewidth]{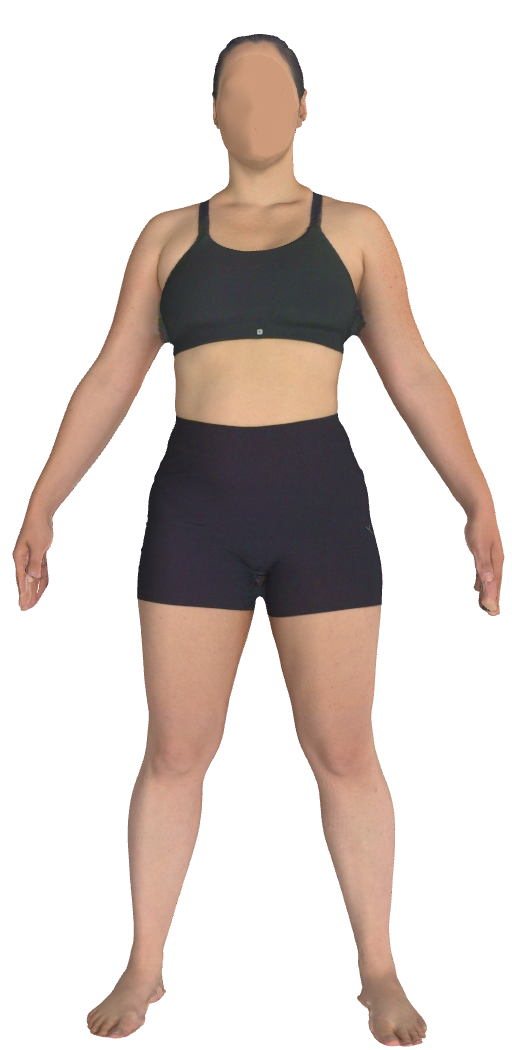}
\caption{$\gls{mesh}_t$}
\label{skin3d:fig:source}
\end{subfigure}
\begin{subfigure}[b]{0.24\linewidth}
\includegraphics[width=\linewidth]{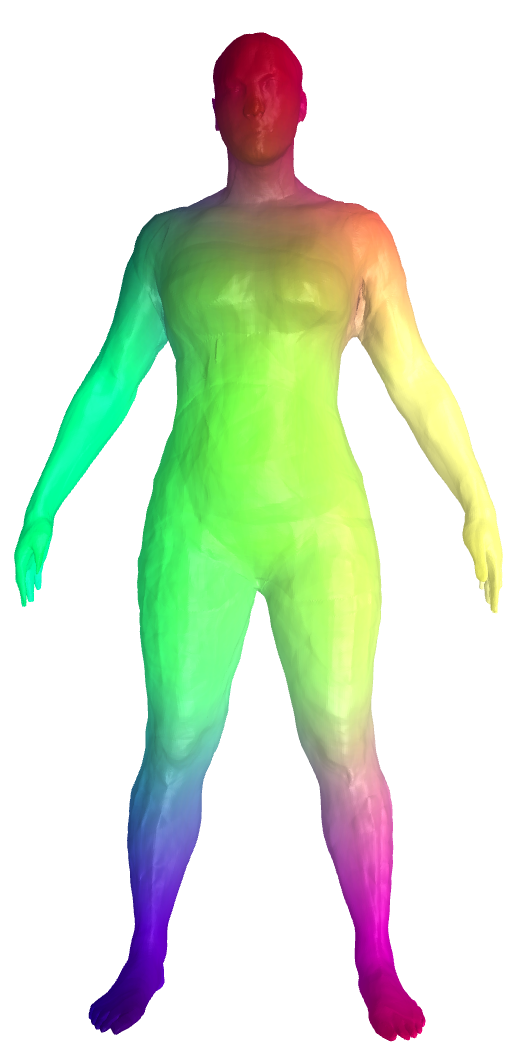}
\caption{$R_t$}
\label{skin3d:fig:source-reconstructed}
\end{subfigure}
\begin{subfigure}[b]{0.24\linewidth}
\includegraphics[width=\linewidth]{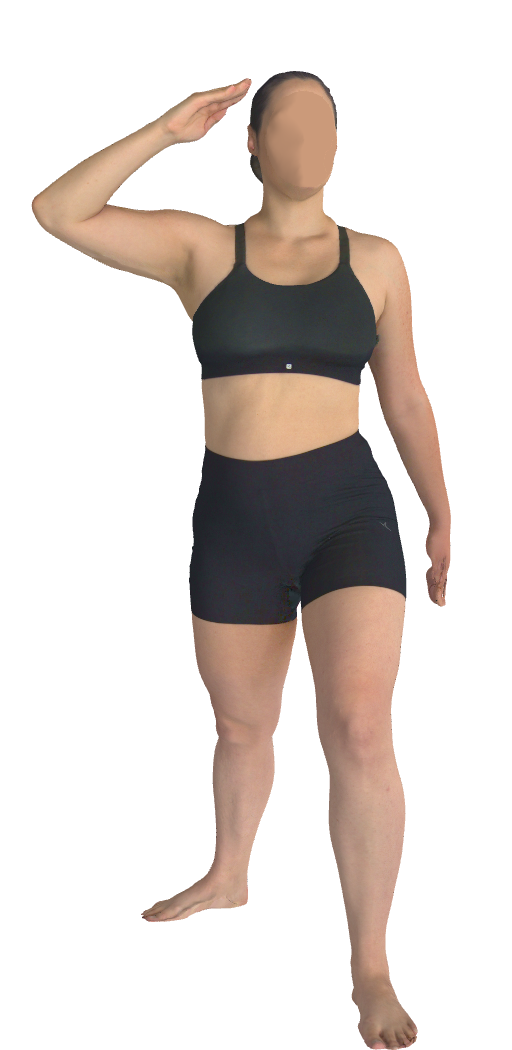}
\caption{$\gls{mesh}_{t+1}$}
\label{skin3d:fig:target}
\end{subfigure}
\begin{subfigure}[b]{0.24\linewidth}
\includegraphics[width=\linewidth]{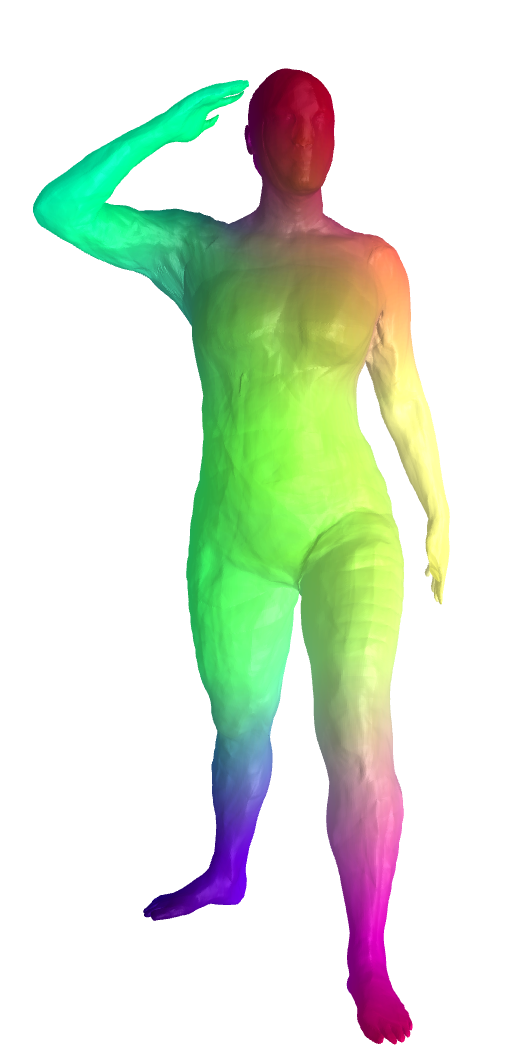}
\caption{$R_{t+1}$}
\label{skin3d:fig:target-reconstructed}
\end{subfigure}
\caption{Correspondence between two 3D meshes. (a) and (c) show meshes of the same subject in different poses. (b) and (d) show the reconstructed meshes, where the colours indicate vertex correspondence, obtained using 3D-CODED~\citep{Groueix2018}.}
\label{skin3d:fig:bodytex-reconstructed}
\end{figure}

The same patient can be scanned at $T$ different times to form a set of 3D meshes $\{\gls{mesh}_t\}_{t=1}^T$. Our goal is to track the lesions across time, which requires us to account for variations in the subject across scans (\eg pose may slightly vary across scans even if the subject is instructed to conform to a standard body position). As the 3D spatial coordinates alone are unsuitable to directly determine anatomical correspondence across meshes, we rely on 3D-CODED, a shape deformation network that uses deep learning to match deformable shapes~\citep{Groueix2018,Deprelle2019}, to determine mesh correspondence. We use the human template, trained network, and default optimization parameters as provided by 3D-CODED, to determine anatomical correspondences of the vertices across scans of the same subject. The full details of the 3D-CODED approach are provided in~\citet{Groueix2018}, but here we highlight that 3D-CODED outputs reconstructed vertices $\gls{reconVertices} \in \gls{realNumbers}^{N_A \times 3}$ that map to vertices in a common template $\gls{templateVertices} \in \gls{realNumbers}^{N_A \times 3}$ (where $N_A$ are the number of vertices within the template), which we use to determine anatomical correspondences (Fig.~\ref{skin3d:fig:bodytex-reconstructed}). Specifically, given $\gls{reconVertices}_t$ as the reconstructed vertices of $\gls{mesh}_t$, for the $i$-th vertex $\gls{vertex}^{(i)}_t$ in \gls{mesh}$_t$, we find the index $j^*$ of the closest reconstructed vertex,
\begin{equation}
   j^* = \argmin_{j \in \{1, \dots, N_A \}} L_2(\gls{reconVertex}_t^{(j)}, \gls{vertex}^{(i)}_t)
   \label{skin3d:eqn:closestReconstructed}
\end{equation}
where $\gls{reconVertex}_t^{(j)} \in \gls{reconVertices}_t$ is the $j$-th reconstructed vertex; and $L_2(\cdot)$ computes the $L_2$ distance between the reconstructed and the original vertex. As the reconstructed vertices $\gls{reconVertices}_t$ and $\gls{reconVertices}_{t+1}$ share the common template \gls{templateVertices}, the $j^*$-th vertex index gives us the anatomically corresponding vertex in $\gls{reconVertices}_{t+1}$ (\ie $\gls{reconVertex}^{(j)}_t \in \gls{reconVertices}_t$ and $\gls{reconVertex}^{(j)}_{t+1} \in \gls{reconVertices}_{t+1}$ can have different spatial coordinates, but point to the same anatomical location). Thus, we find the anatomically corresponding point in $\gls{mesh}_{t+1}$ by,
\begin{equation}
   k^* = \argmin_{k \in \{1, \dots, N_{t+1}\} } L_2(\gls{vertex}_{t+1}^{(k)}, r^{(j^*)}_{t+1})
   \label{skin3d:eqn:closestTarget}
\end{equation}
where $N_{t+1}$ is the number of vertices in \gls{mesh}$_{t+1}$; the vertex $\gls{vertex}^{(k)}_{t+1} \in \gls{mesh}_{t+1}$; and $\gls{reconVertex}^{(j^*)}_{t+1} \in \gls{reconVertices}_{t+1}$ is the vertex reconstructed from $\gls{mesh}_{t+1}$ with the $j^*$-th index as determined by Eq.~\ref{skin3d:eqn:closestReconstructed}. With the $k^*$-th index, we then determine that $\gls{vertex}^{(k^*)}_{t+1} \in \gls{mesh}_{t+1}$ corresponds anatomically to $\gls{vertex}^{(i)}_t \in \gls{mesh}_t$, which allows us to map anatomically corresponding vertices in $\gls{mesh}_k$ to vertices in $\gls{mesh}_{k+1}$.

\subsection{Tracking Lesions Across Poses and Time}
\label{skin3d:sec:tracking}
Given a pair of meshes, $\gls{mesh}_t$ and $\gls{mesh}_{t+1}$, of the same subject imaged at different times, and the detected lesions, $\gls{predLesions2d}_{t}$ and $\gls{predLesions2d}_{t+1}$, detected on the 2D texture images (Eq.~\ref{skin3d:eqn:lesionDetector}), our goal is to determine lesion correspondence. One approach, as proposed by~\citet{Bogo2014a}, is to register the meshes to a mesh template with a common UV mapping to form a standardized texture image with semantically aligned pixels, and compare the resulting pixel locations in the texture images to identify corresponding lesions. However, a drawback of this approach is that anatomically close 3D locations in the mesh may have large 2D distances in the texture image due to the UV mapping. This may be especially problematic in the cases when lesions occur near a seam (\eg Fig.~\ref{skin3d:fig:bodytex-detect2d}), where a small inaccuracy in the mesh registration process may place the lesions on the opposite side of the seam to be UV mapped to a far 2D location and create a large 2D distance. Thus, in this work, we perform our tracking based on 3D geodesic distances between vertices of registered meshes.

We map the center points of the 2D detected lesions to the closest vertices on the corresponding 3D meshes (Eq.~\ref{skin3d:eqn:mapped3d}). We denote these 3D vertices that correspond to lesion center points as $\gls{predLesions3d}_t \in \gls{vertices}_t$, where $\gls{predLesions3d}_t$ is a subset of the vertices found within the original mesh's vertices $\gls{vertices}_t$ (\ie $\gls{predLesion3d}^{(i)}_t \in \gls{predLesions3d}_t$ is the 3D coordinates of the $i$-th lesion $\gls{predLesion2d}_t^{(i)}$ detected on the 2D texture image). 

We seek to establish correspondence between the lesions $\gls{predLesion3d}_t\in \gls{predLesions3d}_t$ and $\gls{predLesion3d}_{t+1}\in \gls{predLesions3d}_{t+1}$ of any two (temporally adjacent) meshes $\gls{mesh}_t$ and $\gls{mesh}_{t+1}$. The correspondence would allow for tracking changes (\eg size and appearance) in a particular lesion across time (similar to previous work on 2D skin images~\citep{Mirzaalian2016}). To this end, we optimize the matching function,
\begin{equation}
{\gls{3dmatching}} :(\gls{predLesion3d}_t, \gls{predLesion3d}_{t+1}) \longrightarrow \{0, 1\}
\end{equation}
which gives 1, when its input arguments, a lesion $\gls{predLesion3d}_t \in \gls{predLesions3d}_{t}$ from $\gls{mesh}_t$ and $\gls{predLesion3d}_{t+1} \in \gls{predLesions3d}_{t+1}$ from $\gls{mesh}_{t+1}$, are set to  correspond, and 0 otherwise. We enforce $\sum_{\gls{predLesion3d}_{t+1}}{\gls{3dmatching}}(\gls{predLesion3d},\gls{predLesion3d}_{t+1}) = 1$ so that a lesion in $\gls{mesh}_t$ is mapped to a single lesion in $\gls{mesh}_{t+1}$. 

We find an optimal $\gls{3dmatching}$ that minimizes a loss function $\gls{3d_match_loss}$, i.e.,

\begin{equation}
   \gls{3dmatching}^* = \argmin_{\gls{3dmatching}} \gls{3d_match_loss}
\end{equation}

\noindent where $\gls{3d_match_loss}$ is designed to encourage that (i) corresponding lesions $\gls{predLesion3d}_t$ in $\gls{mesh}_t$ and $\gls{predLesion3d}_{t+1}$ in $\gls{mesh}_{t+1}$ be located at the same (or more proximate) anatomical locations, which is implemented as a unary term $\gls{unary}$ in $\gls{3d_match_loss}$; and (ii) distances between any pair of lesions, $\gls{predLesion3d}^{(i)}_t$ and $\gls{predLesion3d}^{(j)}_t$, in $\gls{mesh}_t$ be identical (or more similar) to the distances between the corresponding  pair of lesions, $\gls{predLesion3d}^{(m)}_{t+1}$ and $\gls{predLesion3d}^{(n)}_{t+1}$, in $\gls{mesh}_{t+1}$, which is implemented via a binary term $\gls{binary}$ in $\gls{3d_match_loss}$.  Consequently, we write: 

\begin{equation}
    \gls{3d_match_loss}=\alpha \gls{unary}+(1-\alpha)\gls{binary},~~~\alpha\in [0, 1]
\label{eqn:matchingcost}
\end{equation}

\begin{equation}
\gls{unary}(\gls{3dmatching}, \gls{predLesions3d}_{t}, \gls{predLesions3d}_{t+1}) 
= \sum_{\gls{predLesion3d}_t \in \gls{predLesions3d}_{t}, \gls{predLesion3d}_{t+1} \in \gls{predLesions3d}_{t+1}}
\gls{3dmatching}(\gls{predLesion3d}_t,\gls{predLesion3d}_{t+1})\cdot \gls{nodedistance}(\gls{predLesion3d}_{t},  \gls{predLesion3d}_{t+1})
\label{eqn:unary}
\end{equation}

\begin{equation}
\label{eqn:binary}
\begin{aligned}
    \gls{binary}(\gls{3dmatching}&, \gls{predLesions3d}_{t}, \gls{predLesions3d}_{t+1}) 
    = 
    \\
    &
    \!\sum_{
    \mathclap{
        \substack{
    \gls{predLesion3d}_t^{(i)},\gls{predLesion3d}_t^{(j)}\in \gls{predLesions3d}_{t}\\ 
    \gls{predLesion3d}_{t+1}^{(m)}, \gls{predLesion3d}_{t+1}^{(n)}\in \gls{predLesions3d}_{t+1}\\
            }
      }
    } 
    \!\gls{3dmatching}(\gls{predLesion3d}_t^{(i)}\!, \gls{predLesion3d}_{t+1}^{(m)})
    \!\cdot\!
    \gls{3dmatching}(\gls{predLesion3d}_t^{(j)}\!, \gls{predLesion3d}_{t+1}^{(n)})
     \!\cdot\!
        |\gls{node_geodesic}(\gls{predLesion3d}_t^{(i)}\!, \gls{predLesion3d}_t^{(j)}) \!-\!
         \gls{node_geodesic}(\gls{predLesion3d}_{t+1}^{(m)}\!, \gls{predLesion3d}_{t+1}^{(n)})|
\end{aligned}
\end{equation}
\noindent where $\gls{node_geodesic}$ is the geodesic distance calculated on each 3D mesh using fast marching method~\citep{MatlabOTB}; $|\cdot|$ is the absolute value; and $\gls{nodedistance}(\mu,\eta)$ returns a measure of distance between a pair of vertices $\mu,\eta$ (described below).

\textbf{Optimization:} When $\alpha = 1$, Eq.~\ref{eqn:matchingcost}  contains only the unary term. This can be reformulated as a bipartite graph matching problem and optimized through the Kuhn–Munkres algorithm~\citep{munkres1957algorithms}. When $0 < \alpha <1$, a second order regularization term is included in the matching loss, which we optimized using a tensor-based approach~\citep{duchenne2011tensor}.

\textbf{Selection of $\boldsymbol{\gls{nodedistance}}$:} 
The selection of $g$ could result in different optimization results. One choice of $g$ is to set it as the Euclidean distance between two points in 3D. A second choice is to first determine anatomical correspondences between the two meshes, which anatomically registers the 3D coordinates of the lesions $\gls{predLesions3d}_{t}$ and $\gls{predLesions3d}_{t+1}$, then calculate the geodesic distance for each pair of $\gls{predLesion3d}_t$ and $\gls{predLesion3d}_{t+1}$ on a 3D mesh. Specifically, we compute,
\begin{equation}
    \gls{nodedistance}(\gls{predLesion3d}_t, \gls{predLesion3d}_{t+1})=\gls{node_geodesic}(\gls{3dregistration}(\gls{predLesion3d}_t), \gls{predLesion3d}_{t+1})
\end{equation}
where $\gls{node_geodesic}$ is the geodesic distance on $\gls{predLesions3d}_{t+1}$, and $\gls{3dregistration}(\cdot)$ applies the 3D-CODED~\citep{Groueix2018} process described in Section~\ref{skin3d:sec:3dcoded} to map $\gls{predLesion3d}_t$ to the anatomically corresponding vertex in $\gls{predLesions3d}_{t+1}$.

\textbf{Appearing/disappearing lesions:} The above formulation assumes that the numbers of lesions in $\gls{mesh}_t$ and $\gls{mesh}_{t+1}$ are the same and a one-to-one (bijective) correspondence exists. However, in a clinical scenario, longitudinal meshes may exhibit disappearing and newly appearing (or even splitting and merging) lesions. To accommodate these cases in our matching algorithm, we follow a similar approach for matching lesions (including appearing and disappearing) in 2D skin images as described in~\citet{mirzaalian2009graph, Mirzaalian2016}, but we now extend it to 3D meshes. In particular, we extend the set of lesions to be matched, in both $\gls{mesh}_t$ and $\gls{mesh}_{t+1}$ by an additional element, referred to as a ``dummy node", which represents non-existent lesions. A lesion in an earlier scan (i.e., time $t$), which is matched to this dummy node (non-existent lesion) in a future scan ($t+1$), is declared ``disappearing", whereas when a dummy node of an earlier scan ($t$) is matched to a lesion in a future scan ($t+1$), that lesion is declared ``appearing". A lesion must be matched to a dummy-node when it is ``too costly" to match it to an existing lesion. By setting the cost of matching a lesion to the dummy node to a constant value, we allow the matching algorithm to match a lesion to this dummy node to reduce the matching cost. In particular, we extend the set $\gls{mesh}$ to include one additional entry $\gls{predLesion3d}_{\Phi}$ representing the dummy node, and define $\gls{nodedistance}(\gls{predLesion3d},\gls{predLesion3d}_{\Phi})= \text{const.} = \gls{dummy_u_cost}$ in the unary loss of Eqn. \ref{eqn:unary}, and $\gls{node_geodesic}(\gls{predLesion3d},\gls{predLesion3d}_{\Phi})= \text{const.} = \gls{dummy_b_cost}$ in the binary loss of Eqn. \ref{eqn:binary}.

\section{Evaluation and Discussion}
\label{skin3d:sec:evaluation}
We first outline our evaluation metrics used to measure the performance of our proposed lesion detection approach. Then, to address the lack of 3D skin data with annotated skin lesions, we describe our approach to annotate 3DBodyTex~\citep{Saint2018}. Finally, we describe our Faster R-CNN training process, and present our results on lesion detection and tracking.

\subsection{Evaluation Metrics to Detect Bounding Boxes}
\begin{figure}[htb]
\centering
\includegraphics[width=0.95\linewidth]{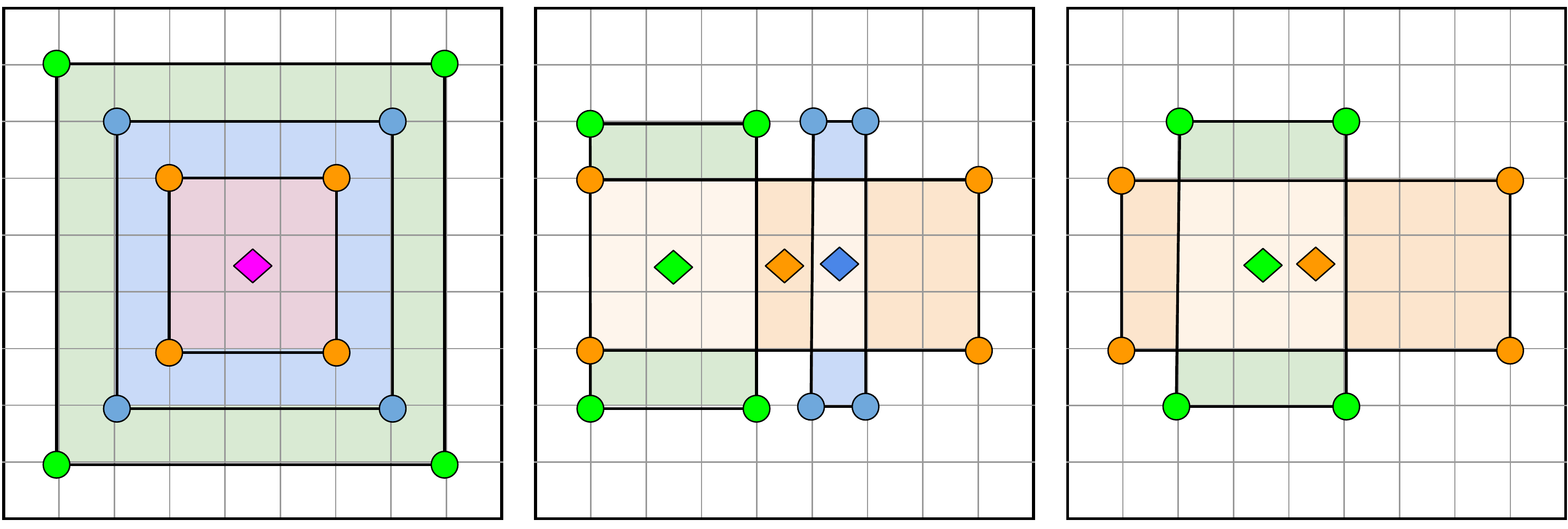}
\caption{(\emph{left}) While the IoU differs among the green, blue, and orange boxes, they have the same centroid (diamond) and are considered as matching using the overlapping centroid metric. (\emph{middle}) The orange box contains the centroid for the green and blue boxes; however, the green and blue boxes do not contain the centroid for the orange box, thus are not considered a match. (\emph{right}) The green and orange boxes match as both contain each other's centroids. Note that the green and orange boxes have the same IoU in the \emph{middle} and \emph{right} figures, but only the \emph{right} figure shows a match using the centroid metric.}
\label{skin3d:fig:centroids}
\end{figure}

\begin{figure}[htb]
\centering
\includegraphics[width=\linewidth]{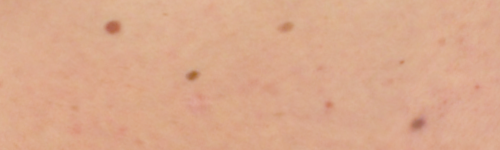}
\includegraphics[width=\linewidth]{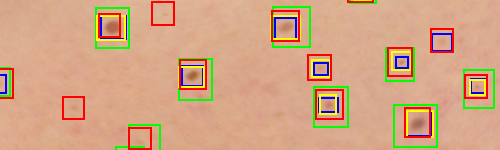}
\caption{The top image shows a zoomed in skin region without any annotations. The bottom image shows the same location but with three different human annotators (blue, green, and yellow boxes) and the machine bounding box predictions (red boxes). The differences among human annotators are shown (\eg larger green boxes vs. the tight blue boxes) motivating our choice to use the centroid matching criteria that is not dependent on an IoU threshold.}
\label{skin3d:fig:bodytex-multi-annotator}
\end{figure}

To measure performance, object detection evaluations typically rely on a intersection over union (IoU) threshold score to determine a ``correct detection'' or a ``match'' between a detected and ground truth bounding box~\citep{Padilla2020}. However, in our pigmented skin lesion case, where the objects of interest occupy a fraction of the entire image, the manual annotations often contain a varying amount of the surrounding skin (Fig.~\ref{skin3d:fig:bodytex-multi-annotator}). As our primary goal is to localize a bounding box that encapsulates the pigmented skin lesion rather than obtaining the precise boundaries around the lesion, we define a metric based on overlapping centroids. Specifically, we define a ``correct detection" or a ``match'' if both the centroids of the manually annotated box \gls{lesion2d} and the model's predicted box \gls{predLesion2d} are enclosed by both,
\begin{equation}
    \left( c(\gls{lesion2d}) \in \gls{predLesion2d} \right) \land \left( c(\gls{predLesion2d}) \in \gls{lesion2d} \right)
    \label{skin3d:eq:centroid}
\end{equation}
where $c(\cdot)$ computes the centroid of the box. Fig.~\ref{skin3d:fig:centroids} shows examples using the overlapping centroid metric on various bounding boxes.

We report \emph{recall}, which indicates the fraction of the manually annotated lesions the model detected; \emph{precision}, which indicates the fraction of all the model's predicted lesions that match with the manually annotated lesions; and, \emph{average precision} (AP), which measures the area under the precision-recall curve for all the predicted confidence thresholds. To compute precision and recall (and in our figures), we use a confidence threshold of 0.5 to determine the predicted bounding boxes, which removes low-confidence lesion predictions. We report results using both the IoU and the overlapping centroid (Eq.~\ref{skin3d:eq:centroid}) approaches, where an IoU threshold of greater than or equal to 0.5 determines a ``correct detection'' or a ``match''.

\subsection{Partitioning and Annotating 3DBodyTex}
\label{skin3d:sec:annoate-bodytex}
3DBodyTex~\citep{Saint2018} contains 400 meshes of 200 unique subjects with 100 females and 100 males, where each subject is acquired in two different poses. From the 200 unique subjects available in 3DBodyTex, we randomly partitioned the subjects into \numberBodyTexTrainSubjects~training, \numberBodyTexValidSubjects~validation, and \numberBodyTexTestSubjects~static testing subjects, where we ensured that the subjects do not overlap across the partitions and that male and female subjects had an even distribution. As 3DBodyTex has two scanned meshes of the same subject, we select \numberBodyTexLongScans~longitudinal testing scans, which are a subset of the same subjects within the \emph{static} testing partition, but in different body poses. We use the \emph{longitudinal} testing partition to evaluate the lesion tracking component of our algorithm (Section~\ref{skin3d:sec:tracking}), where we manually tracked lesions in the \emph{longitudinal} testing partition with the corresponding lesions in the \emph{static} testing partition. 

To evaluate our approach on human subjects, with challenging visual artefacts (\eg hair, cloths), we manually annotate candidate pigmented skin lesions visible on the 2D texture images provided by 3DBodyTex. We consider these manual annotations as ``candidate''  pigmented skin lesions since although these regions appear to contain a pigmented skin lesion, given the scan resolution and the subjectivity of the human annotator, these may contain artefacts that are visually similar to lesions. Furthermore, as we, the annotators, are non-experts (\ie non-dermatologists), these manual annotations are areas of interest that could be noted for the dermatologist to review. 

We choose to annotate the 2D texture image since our target pathology (pigmented skin lesions) is visible within the 2D texture image and is localized within a relatively small region of the skin. This allows us to provide many annotations in 2D without the more cumbersome challenges of 3D annotations, where the 3D body must be navigated in a 3D viewer and parts of the body may not be easily viewed (\eg underneath the arm when the arms are in a downward position as in Fig.~\ref{skin3d:fig:bodytex-detect3d}).

To manually annotate the lesions within the texture images, we used the VGG Image Annotator (VIA)~\citep{Dutta2019,dutta2016via} to place bounding boxes around ``candidate'' lesions. To help standardize the annotations, the texture images were shown zoomed in between five and six times using VIA~\citep{dutta2016via}, and the human annotators were instructed to annotate regions that visually appeared to the annotator as being a pigmented skin lesion. As the precise boundary of the lesion is difficult to determine and our primary goal is to localize the lesions, we annotated with the goal of obtaining a lesion within the center of the bounding box, where some subjectivity to the size of the bounding box was allowed. To better account for differences among human annotations, three human annotators (co-authors MZ, SS, KA) each annotated (Fig.~\ref{skin3d:fig:bodytex-multi-annotator}) the \emph{static} testing partition (used in Section~\ref{skin3d:sec:3dbodytex-static}) to allow us to compare against multiple source of ``ground truth" and measure the performance among the annotators. To evaluate our lesion tracking approach and to form our \emph{longitudinal} testing partition, given two meshes of the same subject $\gls{mesh}_t$ and $\gls{mesh}_{t+1}$ taken at different times and in different poses, we visually determined lesions that correspond (\ie is the same lesion) to each other across the meshes. Specifically, for a lesion annotated by a bounding box on the texture image $\gls{textureImage}_t$, we manually determine the bounding box of the corresponding lesion within $\gls{textureImage}_{t+1}$ and assign both bounding boxes the same unique identifier. This unique identifier assigned to pairs of matching lesions allows us to determine the ground truth lesion correspondence across meshes. Using ten subjects scanned in two different poses, we manually tracked 196 lesions across the pairs of meshes (the \emph{longitudinal} test set used in Section~\ref{skin3d:sec:3dbodytex-tracking}). Due to the effort required to manually track lesions across subjects, we annotated and tracked a subset of the lesions (\ie $19.6 \pm 1.69$ corresponding lesions on each mesh) within the \emph{longitudinal} testing partition. 

We used the high-resolution meshes of 3DBodyTex, where each mesh has $\approx$300K vertices and $\approx$600K faces. From the total 400 meshes, we randomly selected one mesh from each subject and annotated \numberBodyTexScansAnnotated~meshes, resulting in \numberBodyTexTrainScans~training meshes (9 subjects had both meshes annotated), \numberBodyTexValidSubjects~validation meshes, \numberBodyTexTestSubjects~static test meshes, and \numberBodyTexLongScans~longitudinal test meshes. We manually annotated a total of \numberBodyTexLesionsAnnotated~bounding boxes (including texture images annotated multiple times), where the average width and height of the bounding boxes on the 2D texture images were \annotatedBodyTexAvgWidth~and \annotatedBodyTexAvgHeight~pixels respectively.

\subsection{Training~\fasterrcnn~on 3DBodyTex}
\label{skin3d:sec:rcnn}

We trained a \fasterrcnn~\citep{Ren2016} to automatically predict the lesion bounding boxes. The Faster R-CNN is composed of a ResNet-50 network~\citep{he2016deep} pretrained on the COCO dataset~\citep{lin2014microsoft}. For each spatial location, we generate 4 \fasterrcnn~anchor boxes of scale \{32, 24, 16, 8\} with aspect ratio of \{0.75, 1, 1.25\}. We optimized the network using Adam~\citep{Kingma2015} and trained over the 3DBodyTex texture images, using an initial learning rate of 0.001 for 20 epochs, followed by a learning rate of 0.0001 for 10 epochs. 

To reduce GPU memory requirements, we divide the texture image (4096 $\times$ 4096 pixels) into 16 non-overlapping sub-images (1024 $\times$ 1024 pixels) and train using a batch size of 12 sub-images. We note that due to the small size of the lesions, we do not downsample the original texture images as the resulting downsampled images reduce the visual appearance of lesions, preventing many of the lesions from being visible and hence detected. We select the model weights that had the highest average precision score on the validation partition to evaluate the performance on the testing partition. We use non-maximum suppression with an IoU threshold of 0.01 to keep the highest scoring predicted bounding box when bounding boxes overlap.

\subsection{Detecting Lesions on 3DBodyTex Textured Images}
\label{skin3d:sec:3dbodytex-static}

Table~\ref{skin3d:tab:bodytex_results} shows the performance over the static testing partition when considering bounding boxes from the three different human annotators (A1, A2, A3) as the source of ``ground truth''. The reported metrics are improved using the centroid metric as overlapping bounding boxes that delineated different amounts of healthy skin were often considered incorrect using the IoU threshold matching criteria. We show an example of the variability of the human annotations in Fig.~\ref{skin3d:fig:bodytex-multi-annotator}. 

\begin{table}[htb]
    \caption{The performance on the static test partition when automatically detecting pigmented skin lesions within the 3DBodyTex dataset, where three different human annotators (A1, A2, A3) are considered as ``ground truth''. The results shown are averaged over each scan with the standard deviation among each scan shown in brackets. \emph{Avg. Precision} indicates the average precision.}
    \centering
    \begin{tabular}{@{}c c c c c@{}}
    GT & Match & Precision & Recall & Avg. Precision \\
    \toprule
    \multirow{2}{*}{A1} 
    & IoU & 0.21 (0.11) & 0.17 (0.08) & 0.08 (0.06) \\
    & Centroid & 0.79 (0.12) & 0.67 (0.13) & 0.75 (0.12) \\

    \hline
    \multirow{2}{*}{A2} 
    & IoU & 0.50 (0.17) & 0.72 (0.18) & 0.59 (0.21) \\
    & Centroid & 0.62 (0.17) & 0.88 (0.14) & 0.84 (0.10) \\

    \hline
    \multirow{2}{*}{A3} 
    & IoU & 0.32 (0.13) & 0.73 (0.25) & 0.56 (0.23) \\
    & Centroid & 0.44 (0.15) & 0.90 (0.16) & 0.77 (0.14) \\
    \bottomrule
    \end{tabular}
    \label{skin3d:tab:bodytex_results}
\end{table}

\begin{figure}[htb]
\centering
\begin{subfigure}[b]{0.49\linewidth}
\includegraphics[width=\linewidth]{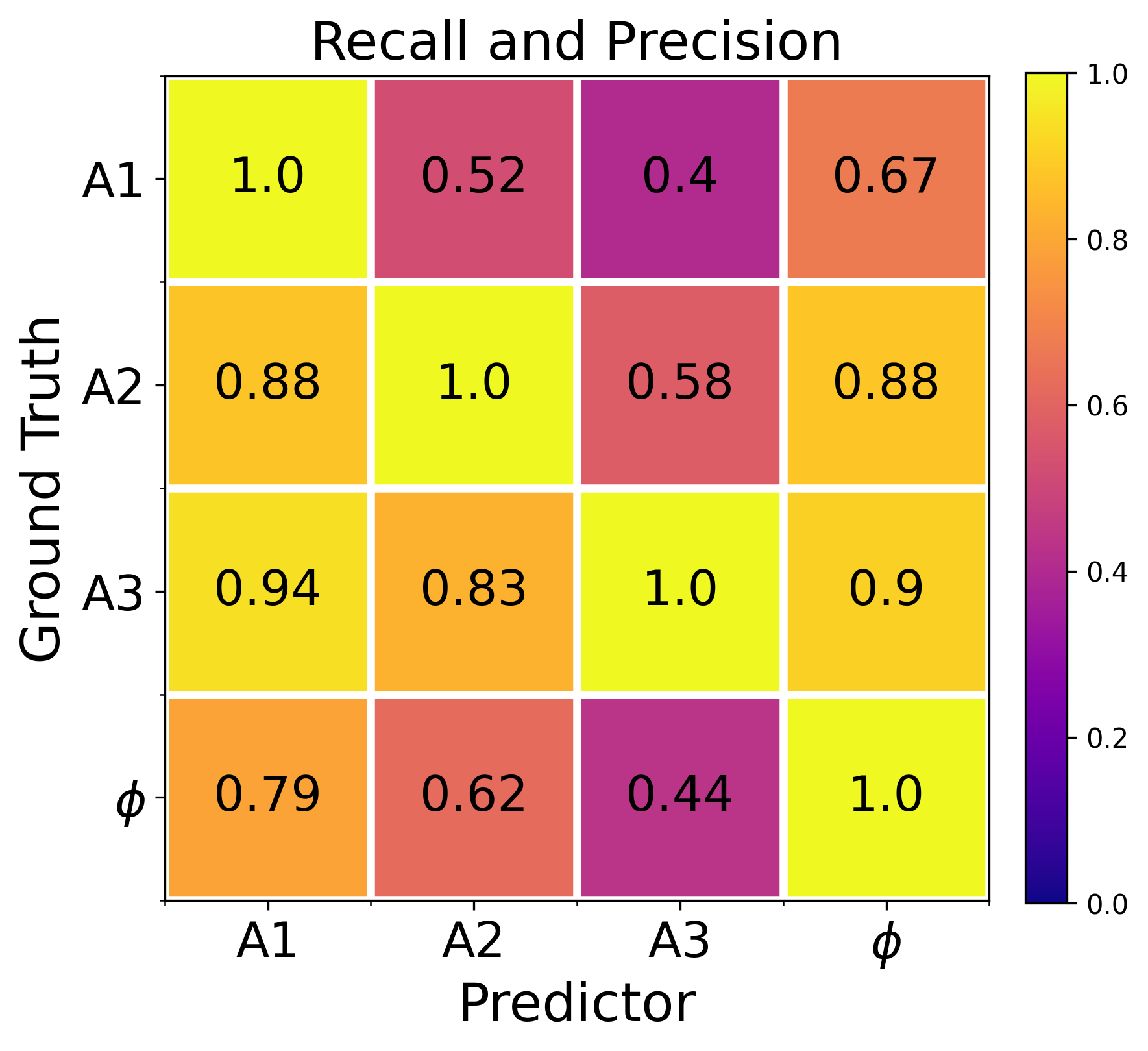}
\caption{}
\label{skin3d:fig:bodytex-multi-annotator-recall-prec}
\end{subfigure}
\begin{subfigure}[b]{0.49\linewidth}
\includegraphics[width=\linewidth]{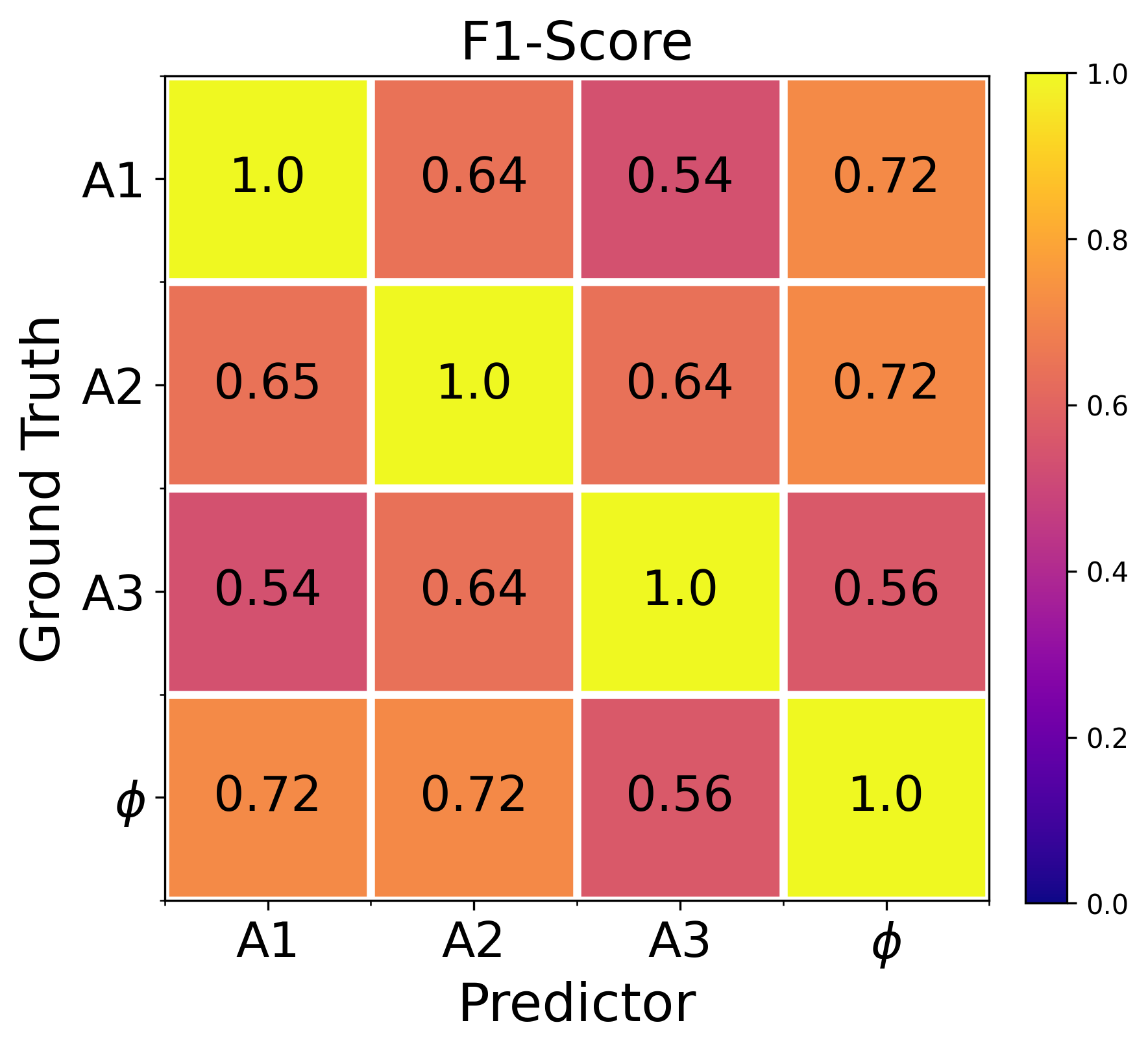}
\caption{}
\label{skin3d:fig:bodytex-multi-annotator-f1}
\end{subfigure}
\caption{Pair-wise performance on the static test set with three human annotators (A1, A2, A3) and the machine annotator ($\phi$). The y-axis indicates which annotator was used as the ground truth, and the x-axis indicates which the annotator was used as the predictions. Each metric is computed per scan and averaged over all scans. (a) The upper and lower triangular matrix displays the recall and precision, respectively. (b). The F1-score combines recall and precision, and shows that the machine has a higher F1-score with two of the three human annotators than the human annotators have with any of each other.}
\label{skin3d:fig:bodytex-multi-annotator-matrix}
\end{figure}

To further explore the performance of the machine annotations compared with human annotations, we set each annotator as the ``ground truth'' and compare the performance of the other annotators with respect to the ``ground truth'' annotator. Fig.~\ref{skin3d:fig:bodytex-multi-annotator-matrix} shows pair-wise precision, recall and F1-score, where each of the human and machine annotators act as the single source of ground truth and predictions. When considering the F1-score as a measure of agreement among annotators, we see that two out of three human annotators agree more with the machine predictions than each other or the third annotator. This result suggests that the machine annotations perform favourably when considering the inter-rater variability among human annotations. We highlight that for this experiment, we compute the performance over the 2D texture images and not the 3D locations and thus do not use the nearest vertex approximations of the lesion centers as made in Eq.~\ref{skin3d:eqn:closestUV} and Eq.~\ref{skin3d:eqn:mapped3d}. Hence, this variability in the human annotator results is not attributed to the approximation of the 3D locations of the lesion centres.

Additional qualitative results using our proposed method were shown in earlier figures. Fig.~\ref{skin3d:fig:bodytex-detect2d} shows the full texture image and the numerous lesions detected using the manual and automatically detected lesions. Fig.~\ref{skin3d:fig:bodytex-detect3d} shows the 3D mesh using the texture image marked with manually and automatically identified lesions. 

\begin{figure}[htb]
\centering
\begin{subfigure}[b]{0.49\linewidth}
\includegraphics[trim={0 20cm 0 0},clip,width=\linewidth]{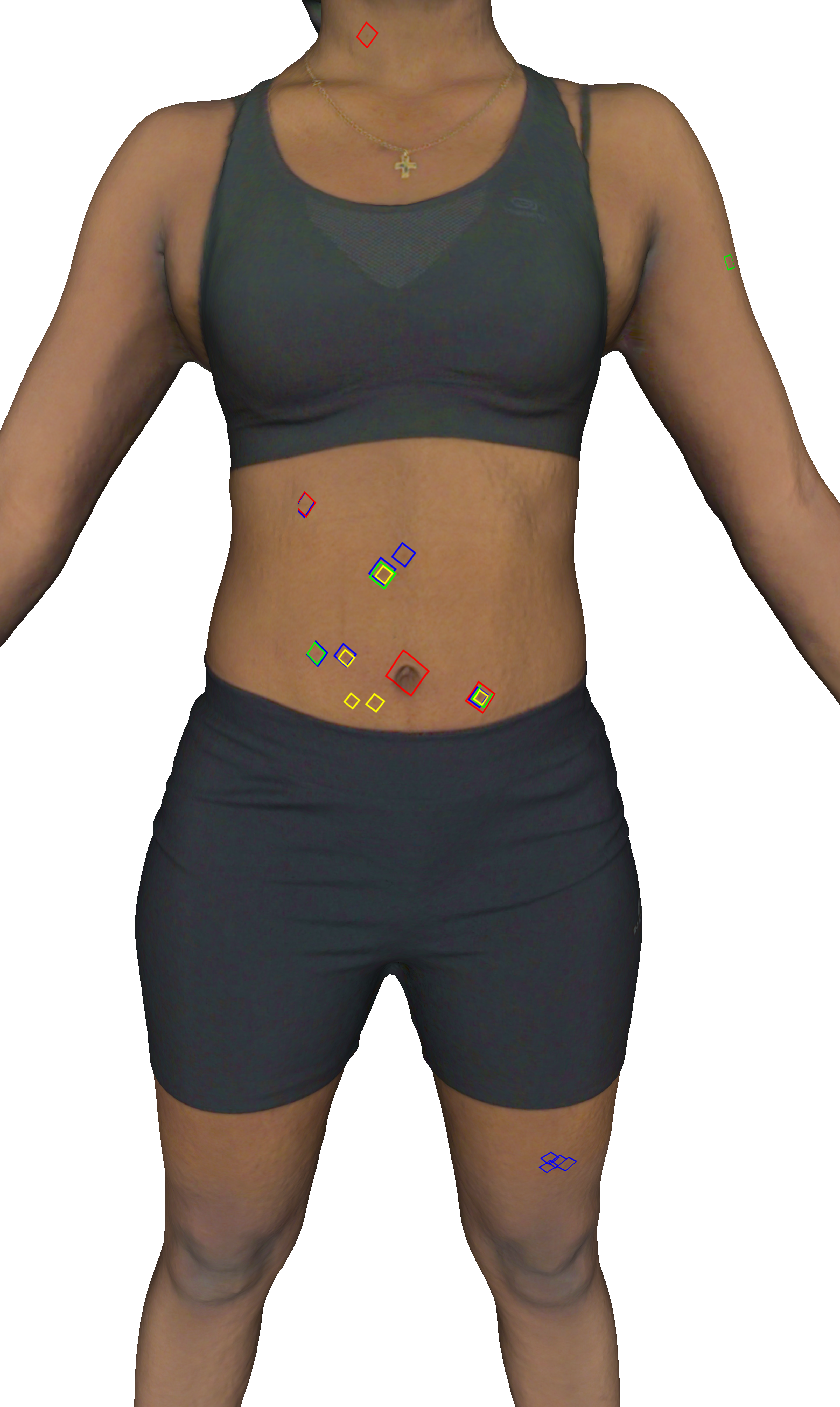}
\caption{}
\end{subfigure}
\begin{subfigure}[b]{0.49\linewidth}
\includegraphics[trim={0 20cm 0 0},clip,width=\linewidth]{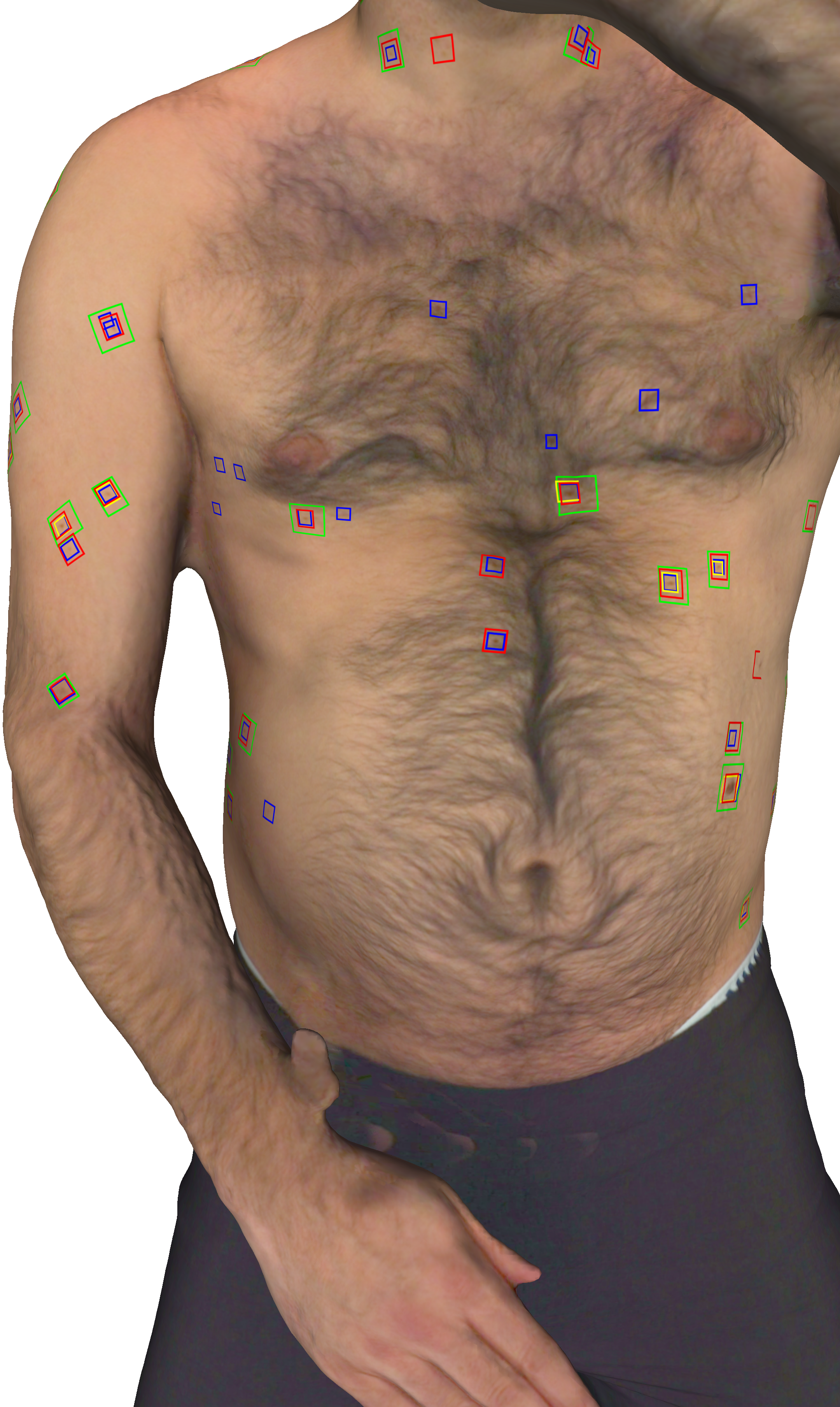}
\caption{}
\end{subfigure}
\caption{Example meshes with the three human annotators (blue, green, and yellow boxes) and the machine (red boxes) bounding box predictions. (a) An example case where the machine incorrectly predicted the navel as a lesion. (b) An example case where the machine has learned to distinguish anatomy from lesions (\eg navel, nipples, hair).}
\label{skin3d:fig:bodytex-multi-example}
\end{figure}

We note that training over all the regions of the body allows our model to learn to distinguish the visual characteristics of salient parts of the anatomy from the lesions to avoid detecting false positive lesions at normal anatomical locations such as the navel and hair. We highlight examples cases in Fig.~\ref{skin3d:fig:bodytex-multi-example}, and note that, in general, our model avoids these types of errors.

\begin{figure*}[htb]
\centering
\includegraphics[width=\linewidth]{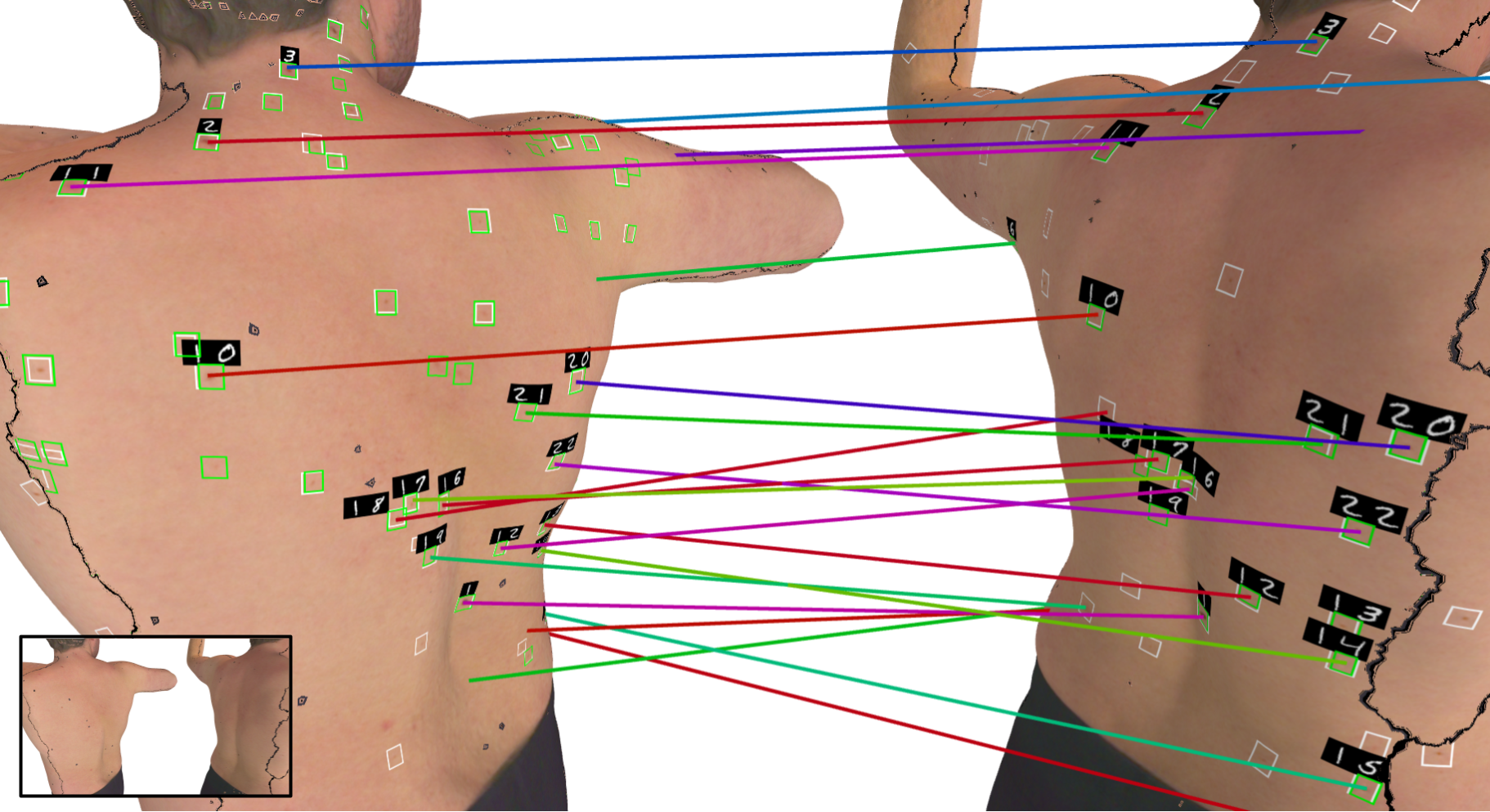}
\caption{Longitudinal tracking for a subset of the annotated lesions, with the unannotated meshes shown in the bottom left. The green boxes indicate the manual annotations. The white boxes indicate the automatic annotations. The coloured lines between the lesions indicate the automatically determined lesion correspondences for the subset of lesions we manually determined correspondences for (as represented by the numbers that indicate corresponding lesions determined manually). The dark curves on the meshes indicate the boundaries of the texture image seams. We note that the scan on the right is part of the longitudinal testing partition, and hence was not densely annotated.}
\label{skin3d:fig:bodytex-long}
\end{figure*}

\subsection{Tracking Lesions on 3DBodyTex}
\label{skin3d:sec:3dbodytex-tracking}

We use the 3DBodyTex \emph{longitudinal} testing partition (Section~\ref{skin3d:sec:annoate-bodytex}) to evaluate the tracking algorithm proposed in Section \ref{skin3d:sec:tracking}. For each mesh belonging to the same subject, we gather the top-$k$ scoring predicted bounding boxes that have scores greater than 0.5 using the trained \fasterrcnn~(Eq.~\ref{skin3d:eqn:lesionDetector}), where $k$ is the minimum number of detected lesions on a mesh of the same subject (we set a maximum value of $k=100$). These detected bounding boxes serve as input to the tracking algorithm.

To measure our performance on correctly tracking the same corresponding lesions across the meshes, we define a \emph{longitudinal lesion} as a manually tracked lesion from the longitudinal testing partition. We report (i) a \emph{matching accuracy} metric that computes `the total number of automatically detected longitudinal lesion pairs correctly tracked' divided by `the total number of automatically detected longitudinal lesions that occur in both scans'; and, (ii) a \emph{longitudinal accuracy} metric defined as `the total number of automatically detected longitudinal lesions pairs correctly tracked' divided by `the total number of pairs within the longitudinal lesion set'. We report the matching accuracy in order to better separate the performance of the tracking approach from the \fasterrcnn~lesion detections, while we report the longitudinal accuracy to capture the effect of accumulating detection errors when using the \fasterrcnn~and  the tracking algorithm together.

As discussed in Section~\ref{skin3d:sec:annoate-bodytex}, the \emph{longitudinal} testing partition contains 20 meshes representing 10 subjects, each with at least 15 and at most 22 pairs of corresponding lesions determined manually, totaling to 196 lesion pair correspondences across 10 subjects with an average of $19.6 \pm 1.69$ lesions per subject. These lesion pairs represent a sample of the total lesions on a subject. In our following experiments, we report results over the entire longitudinal testing partition, and over a subset of the longitudinal testing partition that we refer to as the \emph{prominent lesion pairs}. These prominent lesion pairs are those lesions that are, in general, more easily identifiable to the human eye and are defined as the first 10 pairs that were manually tracked on each subject. We found that, while we could confidently identify and match 10 pairs of lesions per subject, it became considerably more challenging to identify and track subsequent lesion pairs (see Fig.~\ref{skin3d:fig:bodytex-multi-annotator} and Fig.~\ref{skin3d:fig:bodytex-multi-annotator-matrix}, which illustrate the variability among human annotators). This is reflected in Table~\ref{skin3d:tab:full_prominent_compare}, where our \fasterrcnn~achieves a high average recall value of 0.96 when detecting the prominent lesion pairs, and a lower average recall of 0.78 when evaluating over the entire longitudinal testing partition (which is comparable to the recall on the static test set in Table~\ref{skin3d:tab:bodytex_results}). We highlight that, while the prominent lesion pairs are a subset of the entire longitudinal testing partition, the lesion tracking itself is performed on the \emph{exact same set} of predicted lesions. 

Table~\ref{skin3d:tab:bodytex_matching} shows results for the matching and the longitudinal accuracy when including and excluding the unary and binary terms and using Euclidean or geodesic distances, over both the entire longitudinal testing partition (Table~\ref{skin3d:tab:bodytex_matching_full}) and the prominent lesion pairs (Table~\ref{skin3d:tab:bodytex_matching_prominent}), where we empirically set $\gls{dummy_u_cost}=0.5$ and $\gls{dummy_b_cost}=0.5$. We see that the proposed tracking approach (Section~\ref{skin3d:sec:tracking}), which includes both the unary and binary terms computed using geodesic distances, achieves the highest matching and longitudinal accuracy performance. The longitudinal accuracy is consistently lower than the matching accuracy as the longitudinal accuracy metric considers corresponding longitudinal lesions not detected by the \fasterrcnn~as not matched. Table \ref{skin3d:tab:full_prominent_compare} summarizes the performance of the proposed automated lesion tracking system using geodesic distances for the unary and binary terms, where we show results for the entire longitudinal testing partition and the prominent lesion pairs. The performance over the prominent lesions pairs is greater than the entire longitudinal testing partition as the prominent lesion pairs represent a subset of lesions within the longitudinal testing partition more identifiable to the human eye.

Fig.~\ref{skin3d:fig:bodytex-long} shows qualitative results. We highlight that, although the number of pairs of manually matched lesions are limited, due to the manual effort required to determine correct lesion correspondences, we perform our lesion tracking process over a much larger set of detected lesions (on average, 66.6 automatically detected lesions per scan).

\begin{table}[htb]
    \caption{The performance using the manually annotated pairs of lesions of the entire longitudinal test partition and the prominent lesion pairs. Each metric is computed on a pair of meshes, and averaged across paired meshes with the standard deviation shown in brackets. $\alpha$ indicates the weighting between the unary and binary terms used in the tracking equation (Eq.~\ref{eqn:matchingcost}), where $\alpha=1$ uses only the unary term and $\alpha=0.5$ equally weighs both terms. \emph{Distance} indicates the type of distance measure used, where \emph{Euclidean} considers the $L_2$ distances between vertex coordinates while \emph{geodesic} uses the anatomical correspondence between vertices in paired scans to compute geodesic distances. 
    \emph{Match Acc.} and \emph{Long. Acc.} indicate the matching and longitudinal accuracy metrics as defined in the text, respectively.}
    \centering
    \begin{subtable}{1.0\linewidth}
    \centering
    \begin{tabular}{c c c c}
    $\alpha$ & Distance & Match Acc. & Long. Acc. \\
    \toprule
    1 & Euclidean & 0.44 (0.50) &  0.27 (0.50)\\
    0.5 & Euclidean & 0.31 (0.47) & 0.19 (0.47)\\
    1 & Geodesic & 0.75 (0.44) & 0.45 (0.44)\\
    0.5 & Geodesic & \textbf{0.83 (0.38)} & \textbf{0.50 (0.38)}\\
    \bottomrule
    \end{tabular}
    \caption{Entire longitudinal test partition}
    \label{skin3d:tab:bodytex_matching_full}
    \end{subtable}
    \begin{subtable}{1.0\linewidth}
    \centering
    \begin{tabular}{c c c c}
    $\alpha$ & Distance & Match Acc. & Long. Acc. \\
    \toprule
    1 & Euclidean & 0.42 (0.50) &  0.34 (0.50)\\
    0.5 & Euclidean & 0.35 (0.48) & 0.28 (0.48)\\
    1 & Geodesic & 0.77 (0.43) & 0.62 (0.43)\\
    0.5 & Geodesic & \textbf{0.88 (0.33)} & \textbf{0.71 (0.33)}\\
    \bottomrule
    \end{tabular}
    \caption{Prominent lesion pairs}
    \label{skin3d:tab:bodytex_matching_prominent}
    \end{subtable}
    \label{skin3d:tab:bodytex_matching}
\end{table}

\begin{table}[htb]
    \centering
    \caption{Summary of the performance using the proposed tracking system, over the entire longitudinal testing partition and the prominent lesion pairs. \emph{Det. Recall} refers to detection recall of the \fasterrcnn. \emph{Match Acc.} and \emph{Long. Acc.} indicate the matching accuracy and longitudinal accuracy.} 
    \begin{tabular}{c c c c}
    Lesion Set & Det. Recall & Match Acc. & Long. Acc. \\
    \toprule
    Entire & 0.78 (0.12) &0.83 (0.38) &0.50 (0.38)\\
    Prominent & 0.96 (0.06) & 0.88 (0.33) & 0.71 (0.33)\\
    \bottomrule
    \end{tabular}
    \label{skin3d:tab:full_prominent_compare}
\end{table}

\section{Conclusions}
We presented a novel pipeline for the detection and longitudinal tracking of skin lesions from 3D whole-body textured surface scans. The whole-body field-of-view allows for encoding spatial relationships and capturing the global anatomical context of lesions. The longitudinal analysis of body surface scans allows for monitoring and tracking changes in skin lesions across time, which may aid in early detection of melanoma.

Our novel approach of unwrapping the coloured 3D surface body scan into a 2D texture image can go beyond enabling the utilization of 2D \fasterrcnn~to this 3D problem, to allowing leveraging general 2D computer vision techniques, \eg other object detection methods, segmentation techniques, or image enhancement. However, we note that one of the limitations of detecting lesions on the 2D texture image is that, albeit rarely occurring, lesions that appear on the seams of the texture image may be challenging to identify. Future work may explore texture mappings that are split on non-lesion specific areas (\eg cloths) or extending the 2D analysis to seam-aware methods. A limitation of the 3DBodyTex data is the limited number of subjects showing a variety of skin tones, indicating the need to image more diverse subjects. Another important limitation of the 3DBodyTex meshes is that the time between scans for each subject was relatively short, whereas in clinical practice, there would likely be months to years between scans, which would introduce further challenges such as changing body types and skin appearances that are not expressed in this dataset. In addition, for the 3D tracking, we approximate the 3D location of the lesion with the nearest mesh vertex using the high-resolution meshes from 3DBodyTex. However, in the case of low-resolution meshes, where the lesion location is not well captured by the nearest vertex, this approximation may adversely impact the tracking algorithm. In this case, a low-resolution triangular mesh may be refined into a higher resolution mesh (\eg via subdivision) with vertices that better localize the lesion.

Future work would apply our proposed approach onto scans of human subjects with a known range of diseases and incorporate a disease classification step. Future work may also explore training a machine learning model, \eg deep neural network, to establish correspondences~\citep{Yi2018} among lesions across longitudinal scans, where the data to train the model may be created by synthesizing lesions in a scan at time $t$ and then perturbing the lesion (\eg changing its appearance and size) in another scan at time $t+1$. Further, future work should focus on studying skin colour bias under different lighting conditions, which may necessitate collecting datasets of more diverse populations with varying skin colours, as well as designing more advanced skin colour augmentation methods. In addition to collaborations with dermatologists to collect data from diverse subjects under different acquisition equipment and lighting conditions, future dermatology patients’ 3D data acquisition protocols should consider leveraging the standardized pose guidelines proposed for 2D image acquisition for whole body photography~\citep{Halpern2003}.

Given the rapid advancement in 3D scanning technologies in general, we anticipate that acquiring human body scans will become more convenient and popular (\eg via smartphones), and more accurate. This, in turn, is expected to create more demand for developing improved methods for 3D skin surface analysis methods. We hope that the release of the 3DBodyTex manual annotations will help facilitate research in this direction and be of use to the community.

\section*{Acknowledgments}
Partial funding for this work was provided by the Natural Sciences and Engineering Research Council of Canada (NSERC RGPIN-06752). The authors are grateful to Compute Canada and NVIDIA Corporation for providing the computational resources used in this research. The authors thank Zahra Mirikharaji for discussions in the initial stage of the project, Priyanka Chandrashekar for the initial Faster R-CNN experiments, and the anonymous reviewers for their insightful comments to improve this manuscript.

\footnotesize{
\bibliographystyle{IEEEtranN}
\bibliography{IEEEabrv,refs}{}
}
\end{document}